\pdfoutput=1

\documentclass[11pt]{article}

\usepackage[preprint]{acl2023}

\usepackage{times}
\usepackage{latexsym}

\usepackage[T1]{fontenc}

\usepackage[utf8]{inputenc}

\usepackage{microtype}

\usepackage{inconsolata}

\usepackage{graphicx}
\usepackage[scaled=0.85]{beramono}
\usepackage{amsmath}
\usepackage{amssymb}
\usepackage{booktabs}
\usepackage{listings}
\usepackage{siunitx} 
\usepackage{xcolor}
\usepackage{colortbl} 
\usepackage{hyperref}
\usepackage{algorithm}
\usepackage{algpseudocode}

\usepackage{multirow}

\newcommand{\cready}[1]{}

\newcommand\frameworkname{\texttt{PolySQL}}
\newcommand\frameworknamelong{\texttt{PolySQL} (Poly Dialect SQL Benchmark)}

\usepackage{etoolbox}
\usepackage{pgf} 
\definecolor{lowcolor}{HTML}{44ff44} 
\definecolor{midcolor}{HTML}{ffffff}  
\definecolor{highcolor}{HTML}{ff0000}  
 
\title{\frameworkname: Scaling Text-to-SQL Evaluation \\ Across SQL Dialects via Automated Backend Isomorphism
}

\author{\normalsize
\textbf{Yotam Perlitz$^1$, Elad Venezian$^1$, Corentin Royer$^{1,2}$} \\ 
\textbf{Francesco Fusco$^1$, Andrea Giovannini$^1$}
\\ \\
$^1$IBM Research AI \quad $^2$ETH Zürich \\
\texttt{y.perlitz@ibm.com}
}

\begin{document}
\maketitle

\begin{abstract}
SQL dialects vary in syntax, types, and functions across database engines.
Text-to-SQL benchmarks, however, are predominantly SQLite based.
This creates a critical evaluation gap: cross-dialect evaluation reveals weak per-query agreement (Cohen's $\kappa=0.39$), showing that SQLite performance is an unreliable proxy for other dialects.
Yet extending Text-to-SQL evaluations to other dialects remains prohibitively difficult, requiring expensive manual query transpilation or relying on tools that often fail on complex SQL. To close this gap, we introduce \frameworkname, a novel dual-execution method that eliminates the need for query transpilation by comparing normalized execution results. Notably, our approach achieves higher evaluation fidelity than current art with 100\% query coverage. 
Our large scale study reveals a 10.1\% average accuracy drop from SQLite to other dialects and identifies a significant dialect difficulty hierarchy.
We find this degradation stems from logical rather than syntactic errors (61\% vs. 8\%).
We release our framework code, experiments, and leaderboard to enable rigorous dialect-robust evaluation at \url{https://github.com/IBM/polysql}.
\end{abstract}

\begin{figure}[t]
    \centering
    \includegraphics[width=1\linewidth]{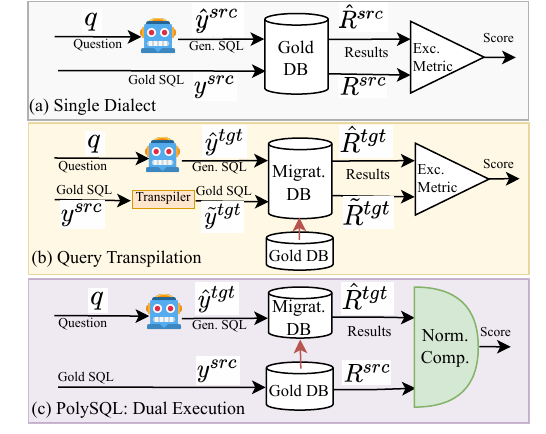}
    \caption{\textbf{Dual Execution vs. Query Transpilation.} 
    \textbf{(a) Single-dialect evaluation.} The model maps an NL question $q$ to SQL, the SQL is executed on the source DB to obtain $\hat{R}^{src}$, that is compared against a gold query result $R^{src}$. \textbf{(b) Transpilation for cross-dialect evaluation.} The model generates an SQL query $\hat{y}^{tgt}$ in the target dialect, which is executed on a migrated (offline) DB . To obtain the reference result $R^{tgt}$, the gold query is first transpiled into the target dialect ($y^{tgt}$) and executed on the same DB; results are compared as in (a). \textbf{(c) \frameworkname.} The gold query is executed on the source DB, while the predicted target-dialect query $\hat{y}^{tgt}$ is executed on a DB migrated (offline) to the target system. A normalized comparator enables result comparison across dialects.
}
    \label{fig:methodology}
\end{figure}

\section{Introduction}

Text-to-SQL systems translate natural language questions into executable SQL queries, enabling non-technical users to interact with DBs. However, SQL is not a monolithic language: different DB engines implement distinct dialects that vary in syntax, type systems, and available functions (see Table~\ref{tab:dialect_divergence} for examples of  dialects syntactic differences). Despite this diversity, Text-to-SQL benchmarks are predominantly evaluated on SQLite, creating a critical gap between research evaluation and production deployment. Researchers and practitioners lack the infrastructure to systematically evaluate how models generalize across dialects. This evaluation gap prevents the community from understanding—and ultimately improving—model behavior when exposed to different dialects.

While multi-dialect benchmarks exist, they remain fundamentally constrained by how dialect variation is introduced. High-fidelity approaches that manually rewrite gold queries for each dialect, such as BIRD \cite{li2023bird}, ensure semantic equivalence but do not scale beyond a handful of dialects. Alternatively, benchmarks that aggregate datasets from diverse execution environments, such as Spider 2.0 \cite{lei2025spider20evaluatinglanguage}, increase surface diversity but conflate dialect differences with changes in question content, making it impossible to isolate the performance degradation caused by syntactic variation alone. Automated transpilation appears to offer a scalable middle ground; however, in practice, both rule-based tools and large language models struggle with complex queries, leading to limited coverage or outright semantic errors. As a result, existing multi-dialect evaluation methodologies fail to provide a reliable, controlled measure of multi-dialect performance.

We introduce \frameworknamelong, a dual-execution framework that eliminates the need for query transpilation (Figure~\ref{fig:methodology}c). Gold queries execute on the source DB (e.g., SQLite) while model predictions execute on migrated target DBs (e.g., PostgreSQL). A normalized comparator bridges formatting differences, enabling direct result comparison across dialects.

This approach leverages a key insight: bridging dialect differences is fundamentally easier at the result level than the query level. While query transpilation requires handling complex dialect-specific syntax, result comparison requires only surface-level normalization (e.g., timestamp formats, numeric precision). DB migration, necessary for both approaches, is handled by mature ELT pipelines\footnote{https://github.com/dlt-hub/dlt}.

To validate our approach, we benchmark against BIRD's manually-transpiled queries. Our dual-execution method outperforms all automated alternatives—including SQLGlot~\cite{sqlglot} static transpilation and LLM-based approaches—while maintaining 100\% query coverage. For this study, we migrate three benchmarks (Spider, BIRD, Archer) to five enterprise dialects (PostgreSQL, MySQL, Snowflake, BigQuery, ClickHouse), generating 15 multi-dialect evaluation environments.

We benchmark 16 models covering a range of sizes and families across all environments to quantify cross-dialect robustness. Our evaluation reveals that SQLite performance is an unreliable predictor, with substantial ranking instability across dialects. Most significantly, we find this degradation is driven by logical errors rather than syntactic ones. Of queries that fail solely due to dialect shift—queries correct on SQLite but incorrect on the target dialect—61\% result in syntactically valid but logically altered SQL, while only 8\% fail due to invalid syntax. Furthermore, our study reveals a clear dialect difficulty hierarchy: PostgreSQL and MySQL form an intermediate tier, while BigQuery and Snowflake constitute the most challenging tier.
Our contributions are as follows:

\begin{enumerate}
\item \textbf{Methodological Contribution:} We implement \frameworkname{}, and establish that DB migration with normalized comparison achieves higher fidelity than other SOTA methods for cross-dialect benchmark porting, while maintaining 100\% execution coverage.

\item \textbf{Empirical Analysis:} Through the largest multi-dialect study to date (16 models, 6 dialects, 3 Benchmarks), we measure a 10.1\% cross-dialect accuracy drop and introduce a Dialect Robustness Score (Appendix~\ref{app:robustness_score}) that measures model consistency across dialects. We demonstrate that degradation is driven by logical errors (61\%) rather than syntactic errors (8\%).

\item \textbf{Resource Release:} We release \frameworkname with extensible support for new dialects and benchmarks, enabling immediate evaluation of arbitrary models. We provide 15 multi-dialect evaluation environments spanning 3 benchmarks and 5 enterprise dialects, and a live leaderboard for community benchmarking\footnote{Code, Results and Environments are shared in the submission zip and would be available upon acceptance}.
\end{enumerate}

\begin{table}[t]
    \centering
    \caption{\textbf{Syntactic Divergence.} A representative sample demonstrates the variation in syntax required to extract the year component for different SQL dialects.
    }
    \label{tab:dialect_divergence}
    \begin{tabular}{l l}
        \toprule
        \textbf{Dialect} & \textbf{ Extract Year Syntax} \\
        \midrule
        SQLite & \texttt{strftime('\%Y', date\_col)} \\
        PostgreSQL & \texttt{EXTRACT(YEAR FROM date\_col)} \\
        MySQL & \texttt{YEAR(date\_col)} \\
        ClickHouse & \texttt{toYear(date\_col)} \\
        \bottomrule
    \end{tabular}
\end{table}

\section{Methodology: \frameworkname{}}
\label{sec:method}

We formulate the problem of Automatic Extension to Multi-Dialect Evaluation as extending single-dialect benchmarks to arbitrary target dialects. Both query transpilation and our dual-execution approach require DB migration, but we eliminate the need for query transpilation itself by implementing a Dual-Execution Protocol: gold queries execute on the source DB while predictions execute on migrated target environments, with results compared via a normalized comparator.

In this section, we define the formal task (\S\ref{ssec:formulation}), describe the automated migration pipeline (\S\ref{ssec:migration}), and detail the dual-execution and normalization protocols (\S\ref{ssec:protocol}).

\subsection{Problem Formulation}
\label{ssec:formulation}
We define a Text-to-SQL benchmark relative to a specific execution dialect $d$.
Let $\mathcal{B}^d = \{(q_i, \mathcal{S}_{i}^d, y_{i}^d)\}_{i=1}^N$ be a benchmark where $q_i$ denotes a natural language question, $\mathcal{S}_{i}^d \in \mathbb{S}^d$ represents the concrete \textit{DB state} (schema and data) realized in the backend engine of dialect $d$, and $y_{i}^d$ is the ground-truth SQL query, executable on $\mathcal{S}_{i}^d$.

In the context of standard evaluation, we are typically provided with a single source benchmark $\mathcal{B}^{d_{src}}$ fixed to a specific source dialect $d_{src}$ (e.g., SQLite), such that we only possess ground-truth queries $y^{d_{src}}$.

Our objective is to generalize this evaluation to an arbitrary target dialect $d_{tgt}$.
This necessitates a fundamental first step: the transformation of the physical execution environment via $\Phi(\mathcal{S}^{src}, d_{tgt}) \rightarrow \mathcal{S}^{tgt}$.
Once the target state $\mathcal{S}^{tgt}$ is established, the critical challenge becomes defining the valid ground truth for this new environment. This presents two distinct pathways: \textbf{transpilation-based} (the standard approach) and \textbf{dual-execution} (our contribution).

\paragraph{1. Synthetic Ground Truth (Query Transpilation)}
The standard approach (Figure~\ref{fig:methodology}b) attempts to map the source gold query to the target dialect via a function $\Psi_{tgt}(y^{src}) \rightarrow \tilde{y}^{tgt}$. Evaluation then proceeds by comparing the model's prediction $\hat{y}^{tgt}$ against this synthetic $\tilde{y}^{tgt}$ on the target DB.
However, due to semantic divergence in SQL dialects, $\Psi_{tgt}$ is inherently \textit{lossy}. A failure to perfectly transpile $y^{src}$ results in a corrupted ground truth, penalizing correct models or rewarding incorrect ones.

\paragraph{2. Cross-Environment Comparison (Dual-Execution)}
Alternatively, we propose to discard the need for a target ground truth query entirely (Figure~\ref{fig:methodology}c).
Instead of relying on a potentially corrupted $\tilde{y}^{tgt}$, we utilize the original, human-verified source query $y^{src}_i$ as the anchor of truth.
We introduce a \textbf{Dual-Execution Protocol} that operates as follows. First, we execute the gold query $y^{src}_i$ on the source DB $\mathcal{S}^{src}_i$ to obtain result set $R^{src}_i$. Second, we execute the model's predicted query $\hat{y}^{tgt}_i$ on the target DB $\mathcal{S}^{tgt}_i$ to obtain result set $R^{tgt}_i$. The approach then evaluates correctness by directly comparing these result sets: $R^{src}_i \approx R^{tgt}_i$, where $\approx$ denotes normalized comparison (detailed in §\ref{ssec:protocol}). For notational simplicity, we henceforth drop the subscript $i$ when discussing a single instance.


In this work, we adopt the second approach.
By decoupling the evaluation metric from query transpilation, we eliminate the noise introduced by static analysis tools, ensuring that the benchmark's fidelity relies solely on the deterministic migration of the DB state.

\subsection{Automated Environment Isomorphism}
\label{ssec:migration}

To implement $\Phi$, we employ the Data Load Tool (dlt)\footnote{\url{https://github.com/dlt-hub/dlt}}, an industrial-grade ELT framework that provides production-ready connectors for major SQL engines. Our migration pipeline operates in three phases: (1) \textit{schema introspection} extracts table definitions, column types, and relationships from the source DB; (2) \textit{type mapping} converts source types to target-compatible equivalents while preserving data semantics (e.g., SQLite's TEXT dates to PostgreSQL's TIMESTAMP); and (3) \textit{bulk data transfer} loads rows into the target engine using optimized batch operations.

Critically, we selectively relax referential integrity constraints during migration. Academic benchmarks frequently contain "dirty" data that violates foreign key constraints—a common property of real-world evaluation datasets. Enforcing strict integrity would cause the target DB to reject these rows, corrupting the benchmark. Our pipeline preserves data fidelity through strictly typed yet constraint-permissive loading.

\subsection{Dual-Execution and Normalized Comparator}
\label{ssec:protocol}

The core challenge of Dual-Execution is \textit{Output Representation Divergence}. While the DB states are isomorphic, the execution engines are heterogeneous. Different backends return semantically equivalent data in structurally distinct formats—for instance, SQLite yields dates as strings and loose numerics, whereas PostgreSQL enforces strict \texttt{TIMESTAMPTZ} objects and high-precision \texttt{DECIMAL} types. Consequently, a naive equality check between $R^{src}$ and $R^{tgt}$ fails, yielding false negatives even for semantically correct queries.

To bridge this gap, we implement a \textbf{Normalized Comparator}. For each example, we execute the gold query on the source DB ($y^{src}$ on $\mathcal{S}^{src}$) and the predicted query on the target DB ($p$ on $\mathcal{S}^{tgt}$). Both executions produce result sets ($R^{src}$ and $R^{tgt}$) that are compared after normalization. This ensures that our evaluation isolates the \textit{semantic correctness} of the model's logic from the \textit{implementation details} of the underlying engine.
The normalization function enforces three key transformations:

\textbf{(1) Bag Semantics Alignment:} For queries without explicit \texttt{ORDER BY}, we canonically sort both result sets to prevent false negatives from engine-specific default ordering.

\textbf{(2) Type Homogenization:} We apply strictly defined casting rules to align driver-specific type representations (e.g., SQLite's loose numerics vs. PostgreSQL's strict \texttt{DECIMAL} types).

\textbf{(3) Precision Tolerance:} We employ $\epsilon=10^{-5}$ tolerance for floating-point comparisons and whitespace normalization to distinguish true semantic errors from arithmetic or formatting artifacts. Full normalization specifications are detailed in Appendix~\ref{app:normalization}.

\begin{table}[t]
    \centering
    \caption{\textbf{SQLite as a Proxy for Target Dialects.} We evaluate whether SQLite performance reliably predicts enterprise dialect performance by comparing each of 16 models on the same 300 evaluation instances (100 per benchmark) across SQLite and five target dialects. Each row shows the agreement between SQLite and a target dialect across all model predictions. Despite moderate linear correlation ($r \approx 0.75$), the poor ranking fidelity ($\rho \approx 0.64$) and low per-query agreement ($\kappa \approx 0.39$) make SQLite an unreliable predictor for deployment.}
    \label{tab:naive_proxy_failure}
    \setlength{\tabcolsep}{4pt} 
    \begin{tabular}{l|ccc}
    \toprule
    \textbf{Dialect} & \textbf{Cohen's $\kappa$} & \textbf{Spear. $\rho$} & \textbf{Pears. $r$} \\
    \midrule
    Snowflake & 0.31 & 0.49 & 0.58 \\
    MySQL & 0.35 & 0.76 & 0.77 \\
    BigQuery & 0.40 & 0.72 & 0.73 \\
    ClickHouse & 0.45 & 0.67 & 0.84 \\
    PostgreSQL & 0.45 & 0.56 & 0.82 \\
    \midrule
    \textit{Average} & \textit{0.39} & \textit{0.64} & \textit{0.75} \\
    \bottomrule
    \end{tabular}
\end{table}

\begin{table*}[h]
    \centering
    \caption{\textbf{Fidelity of Cross-Dialect Evaluation Methods.} Comparison of automated evaluation methods against BIRD’s manually transpiled queries. \textbf{Coverage} denotes the fraction of queries that execute successfully; \textbf{Reliability} reports agreement with the reference using Cohen’s $\kappa$, Spearman’s $\rho$, and Pearson’s $r$. \textit{Pred $\to$ Source (SQLGlot)} transpiles model predictions back to SQLite; \textit{Gold $\to$ Target (SQLGlot)} transpiles gold queries to the target dialect. \textit{LLM-as-a-Judge} evaluates correctness without execution, while \textit{Gold $\to$ Target (LLM)} uses an LLM to transpile gold queries. \frameworkname{} executes predictions on the migrated target DB and gold queries on the source DB, avoiding query transpilation.}
    \label{tab:reliability}
    \setlength{\tabcolsep}{5pt}
    \begin{tabular}{l l | cc | cccc}
    \toprule
    \textbf{Class} & \textbf{Method} & \multicolumn{2}{c|}{\textbf{Requires Migration}} & \multicolumn{4}{c}{\textbf{Reliability \& Ranking}} \\
    \cmidrule(lr){3-4} \cmidrule(lr){5-8}
     &  & \textbf{DBs} & \textbf{Queries} & \textbf{Coverage} & \textbf{$\kappa$} & \textbf{$\rho$} & \textbf{$r$} \\
    \midrule
    Static & Pred $\to$ Source (SQLGlot) & No & Yes & 46.1\% & 0.39 & 0.56 & 0.55 \\
           & Gold $\to$ Target (SQLGlot) & Yes & Yes & 40.9\% & 0.48 & 0.69 & 0.65 \\
    \midrule
    LLM & LLM-as-a-Judge (No Exe) & No & No & 100\% & 0.38 & 0.66 & 0.78 \\
        & Gold $\to$ Target (LLM) & Yes & Yes & 100\% & 0.66 & 0.68 & 0.84 \\
    \midrule
    \textbf{Exec (Ours)} & \textbf{\frameworkname} & Yes & No & 100\% & \textbf{0.72} & \textbf{0.75} & \textbf{0.85} \\
    \bottomrule
    \end{tabular}
    \end{table*}
\section{Experimental Setup}
\label{sec:setup}

This section describes the experimental setup used throughout sections \S\ref{sec:fidelity} and \S\ref{sec:analysis}. We conduct large-scale cross-dialect Text-to-SQL experiments by running SOTA models across multiple SQL dialects and datasets using \frameworkname{} and other baselines as the evaluation infrastructure. In total, we evaluate 16 models over 300 shared evaluation instances across six SQL dialects, resulting in 28,800 executed predictions.

\paragraph{Benchmarks.} We evaluate on three widely used Text-to-SQL benchmarks: Spider \cite{yu2018spider} (cross-domain generalization), BIRD \cite{li2023bird} (DB efficiency and complexity), and Archer \cite{zheng-etal-2024-archer} (arithmetic reasoning). For each benchmark, we randomly sample 100 evaluation instances and migrate all associated DBs from their native SQLite implementations to five enterprise dialects: PostgreSQL, MySQL, Snowflake, BigQuery, and ClickHouse. This yields 15 multi-dialect evaluation environments (3 benchmarks × 5 target dialects). While our experiments use SQLite as the source dialect to align with existing benchmarks, \frameworkname{} is not limited to this setting. Our codebase supports migrations from MySQL, PostgreSQL, and Snowflake as source DBs, demonstrating the framework’s ability to operate bidirectionally across dialects.

\paragraph{Models.} We evaluate 16 state-of-the-art Text-to-SQL models spanning proprietary and open-weights systems, covering a broad range of model sizes and capabilities. The models are grouped into three tiers:
(1) \textbf{Frontier models}: OpenAI GPT-OSS (120B, 20B), Claude~3.5 Sonnet, Claude~3.5 Haiku, and Llama~3.1~405B;
(2) \textbf{High-performance open-weights models}: DeepSeek (V2.5, V3), Qwen~2.5~72B, Llama~3.3~70B, and Llama~4 Maverick;
(3) \textbf{Efficient and code-specialized models}: DeepSeek Coder~33B, Llama~4 Scout~17B, Llama~3.1~8B, Qwen~3~8B, Mistral Small, and Granite~3.3~8B.
All models are evaluated using greedy decoding (temperature~=~0) to ensure reproducibility.

\paragraph{Prompting Strategy.} We adopt a zero-shot-with-instructions setting to evaluate models under explicit dialect constraints without relying on few-shot examples. For each target SQL dialect, we construct a concise set of five dialect-specific syntax guidelines (e.g., preferred date functions or identifier quoting rules), which are injected into the system prompt together with the migrated DB schema (DDL). Guidelines were generated using Gemini-2.5-Pro, manually validated for correctness, and held fixed across all models. Full prompts are provided in the code.

\paragraph{Evaluation.} For each model prediction, we apply a dual-execution evaluation protocol. The model-generated query is executed on the DB migrated to the target dialect, while the corresponding gold query is executed on the original source (SQLite) DB. The resulting outputs are compared using our Normalized Comparator (§\ref{ssec:protocol}), which accounts for dialect-specific formatting differences. This comparison yields a binary correctness signal that is used to compute execution accuracy.

\section{Results: Validating \frameworkname{}}
\label{sec:fidelity}

\noindent To validate the reliability and practicality of \frameworkname{} as a cross-dialect Text-to-SQL evaluation method, we assess how closely automated evaluation approaches align with manually transpiled queries from BIRD, which serve as our reference. Specifically, we compare execution outcomes for all 16 models on BIRD obtained via \frameworkname{} against single-dialect evaluation results derived from BIRD’s human-transpiled queries on the corresponding target dialects. We benchmark our execution-based protocol against four alternatives spanning two classes: (1) rule-based query transpilation using SQLGlot, applied either by transpiling model predictions back to the source dialect or by transpiling gold queries forward to the target dialect; and (2) LLM-based approaches using GPT-oss-120b, either as a judge without execution or as a query transpiler. Agreement is quantified using Cohen’s Kappa, Spearman’s $\rho$, and Pearson’s $r$, capturing alignment at the query level, consistency of model rankings, and correlation of aggregate model performance, respectively. As shown in Table~\ref{tab:reliability}, rule-based transpilation crashes on $54\%$–$59\%$  of queries (coverage $\leq46\%$), while LLM-based approaches achieve full coverage but substantially lower agreement ($\kappa=0.38$–$0.66$). In contrast, our execution-based approach achieves both $100\%$ coverage and the highest reliability ($\kappa=0.72$, $\rho=0.75$), eliminating the need for query transpilation.

\paragraph{The Failure of Static Transpilation.}
Static query transpilation tools (e.g., SQLGlot) exhibit severe \textbf{survivorship bias}: a large fraction of queries fail to transpile due to complex logic or dialect-specific constructs. As a result, reported reliability reflects performance on a reduced and easier subset of the benchmark, inflating agreement metrics. In practical evaluation settings, discarding hard queries undermines both coverage and the validity of model comparison.

\paragraph{The Instability of LLM Judges.}
LLM-based approaches achieve 100\% coverage but trade precision for hallucination. The "LLM-as-a-Judge" baseline achieves a Cohen's Kappa of only $0.38$, indicating that its verdicts are little better than random guessing compared to ground-truth execution. Even when using an LLM to transpile Gold queries ("Gold $\to$ Target"), the reliability ($\kappa=0.66$) lags behind strict execution, as the model frequently generates syntactically valid but semantically incorrect translations.

\paragraph{The \frameworkname{} Solution.}
\frameworkname{} resolves this dilemma by shifting the migration burden from the \textit{query} to the \textit{DB}. By keeping the SQL queries invariant and adapting the schema to support them, we achieve \textbf{Universal Coverage (100\%)} Unlike static tools, we guarantee an execution signal for every sample in the dataset.
Additionally, \textbf{Deterministic Reliability ($\kappa=0.72$)} Unlike LLM judges, our verdicts are based on physical execution, achieving the highest agreement with the source-truth.

This results in the highest ranking fidelity (Spearman $\rho=0.75$) and correlation ($r=0.85$), confirming that \frameworkname{} is the only method capable of benchmarking SOTA models without discarding complex queries or introducing probabilistic noise.

\begin{figure}[t]
    \centering
    \includegraphics[width=1\linewidth]{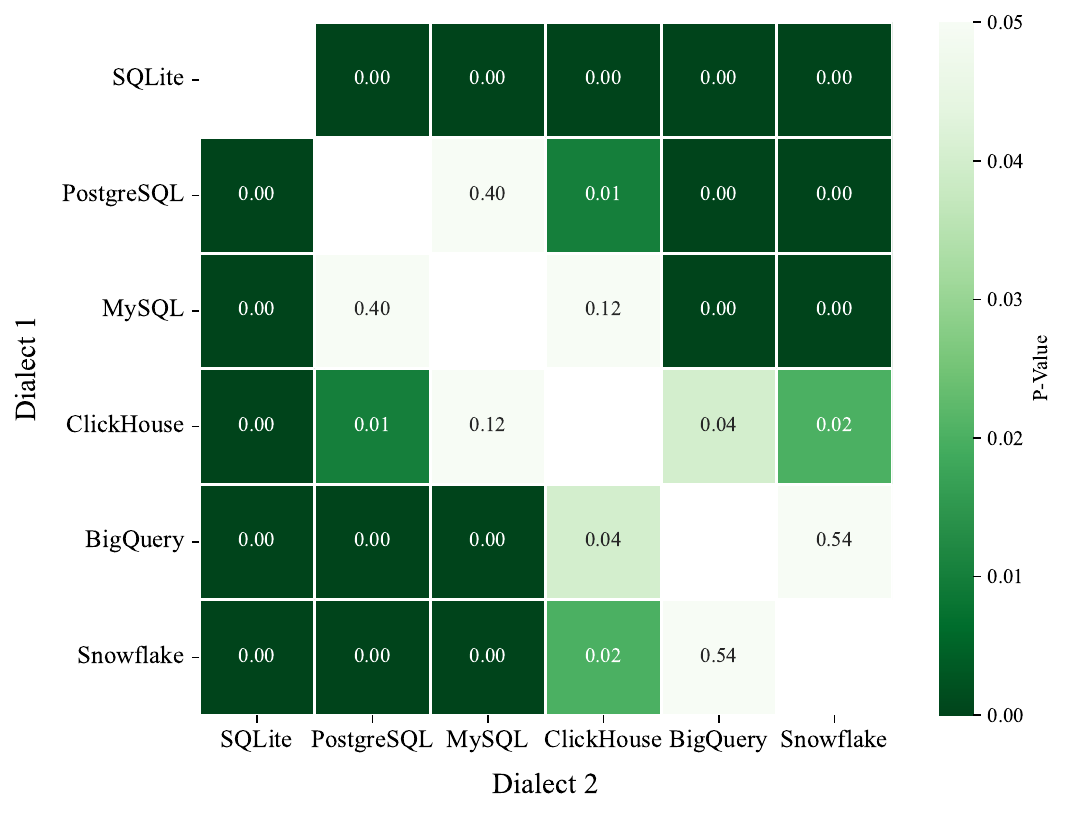}
    \caption{\textbf{Dialect Difficulty Hierarchy via McNemar's Test.} Heatmap of pairwise McNemar's test p-values comparing model performance across dialects. Dark green indicates significant differences ($p<0.05$), while light colors indicate statistical indistinguishability. PostgreSQL and MySQL form a statistically indistinguishable intermediate tier ($p\approx0.40$), while BigQuery and Snowflake constitute the most challenging tier, significantly harder than all others (all $p\approx0.01$). SQLite shows significant differences from all target dialects (all $p<0.00$), confirming the SQLite Premium.}
    \label{fig:dialect_difference_significance}
\end{figure}

\section{Analysis: Multi-dialect Text-to-SQL}
\label{sec:analysis}

\noindent Having validated that \frameworkname{} achieves higher fidelity than query transpilation (§\ref{sec:fidelity}), we now use it to conduct the largest cross-dialect Text-to-SQL study to date. We address three research questions: (1) Can SQLite serve as a reliable proxy for enterprise dialects? (2) How do models rank across different dialects? (3) Why do models fail when transitioning from SQLite to target dialects?

Using the experimental setup described in §\ref{sec:setup} (16 models, 6 dialects, 300 evaluation instances per benchmark), we rigorously quantify the cross-dialect generalization gap. The following analysis demonstrates that SQLite is an insufficient proxy for production dialects (\S\ref{ssec:proxy_validity}), presents the first multi-dialect leaderboard (\S\ref{sec:main_results}), and dissects the root causes of cross-dialect model failure (\S\ref{ssec:cognitive_fragility}).

\subsection{The Necessity of Dialect-Specific Evaluation}
\label{ssec:proxy_validity}

Having validated our framework's fidelity, we now investigate whether SQLite-based evaluation can serve as a reliable proxy for production dialects. We compare each model's performance on SQLite against its performance on each of the five target dialects using the same evaluation instances. The results (Table~\ref{tab:naive_proxy_failure}), measured using the reliability metrics from §\ref{sec:fidelity}, reveal that SQLite exhibits weak per-query agreement (Cohen's $\kappa$=0.39), poor ranking fidelity (Spearman $\rho$=0.64), and only moderate linear correlation (Pearson $r$=0.75) with target dialects, demonstrating a gap between this pragmatic default and production reality.

While a surface-level analysis might suggest validity—indicated by a moderate Pearson correlation ($r \approx 0.75$)—this metric is deceptive. It captures the general trend that "better models are better everywhere" but masks specific failure modes. Crucially, the metrics that define benchmarking utility tell a different story. The Spearman rank correlation ($\rho \approx 0.64$) is significantly lower than the linear correlation, implying that optimizing for a SQLite leaderboard does not guarantee SOTA performance on target dialects.

The failure is most acute when attempting \textbf{fine-grained error analysis}. We employ Cohen's Kappa specifically to test the "fine print" of evaluation: can SQLite reliably identify \textit{which} specific queries fail in a target dialect? The low agreement ($\kappa \approx 0.39$) confirms that it cannot.


\subsection{Multi-Dialect Benchmarking Results} \label{sec:main_results}

\begin{table*}[t]
\centering
\begin{tabular}{llr}
\toprule
\textbf{Category} & \textbf{Represents} & \textbf{Frequency (\%)} \\
\midrule
Schema Linking & Hallucinated tables/columns that don't exist & 9.0\% \\
Filtering/Logic & Wrong row selection, filters, JOINs, or logic & 61.2\% \\
Aggregation/Grouping & Wrong GROUP BY, aggregations, or ordering & 10.9\% \\
Dialect Syntax & Dialect-specific syntax or function errors & 7.7\% \\
Evaluation Framework & Ground truth issues, execution failures, etc. & 11.1\% \\
\bottomrule
\end{tabular}
\caption{Error Classification Categories and Distribution}
\label{tab:error_categories}
\end{table*}
\begin{table*}[t]
\centering
\renewcommand{\arraystretch}{1} 
\setlength{\tabcolsep}{6pt} 
\caption{\textbf{Multi-Dialect Execution Accuracy.} Models are ranked by average performance across 6 dialects. \textbf{Bold} indicates significant ($p<0.05$) best performance per column.}
\label{tab:main_results}

\begin{tabular}{
    @{\extracolsep{2pt}} 
    l 
    S[table-format=2.1, detect-weight] 
    S[table-format=2.1, detect-weight] 
    S[table-format=2.1, detect-weight] 
    S[table-format=2.1, detect-weight] 
    S[table-format=2.1, detect-weight] 
    S[table-format=2.1, detect-weight] 
    >{\columncolor{gray!10}}S[table-format=2.1, detect-weight] 
}
\toprule
\textbf{Model} & {\textbf{SQLite}} & {\textbf{Postgres}} & {\textbf{MySQL}} & {\textbf{Snowflake}} & {\textbf{BigQuery}} & {\textbf{ClickHouse}} & {\textbf{Avg}} \\
\midrule
Claude 3.5 Sonnet & \bfseries 63.1 & \bfseries 50.0 & \bfseries 48.8 & \bfseries 45.2 & \bfseries 47.2 & \bfseries 48.4 & \bfseries 50.5 \\
GPT-OSS-120B      & 54.8 & 42.1 & 43.3 & 41.3 & 40.5 & 40.9 & 43.8 \\
DeepSeek V2.5     & 52.8 & 40.5 & 38.5 & 43.7 & 40.9 & 42.1 & 43.1 \\
DeepSeek V3       & 53.2 & 42.9 & 40.5 & 39.7 & 37.3 & 44.8 & 43.1 \\
Llama 3.1 405B    & 54.8 & 40.9 & 40.5 & 34.5 & 39.3 & 45.2 & 42.5 \\
Claude 3.5 Haiku  & 54.8 & 39.3 & 42.1 & 38.1 & 38.5 & 42.1 & 42.5 \\
GPT-OSS-20B       & 53.2 & 41.7 & 41.7 & 34.5 & 38.5 & 39.7 & 41.5 \\
Llama 4 Maverick  & 51.2 & 42.5 & 38.5 & 38.5 & 36.9 & 39.7 & 41.2 \\
Qwen 2.5 72B      & 52.0 & 39.3 & 38.5 & 35.3 & 36.1 & 43.7 & 40.8 \\
Llama 3.3 70B     & 50.0 & 39.3 & 40.5 & 40.5 & 34.5 & 38.1 & 40.5 \\
Mistral Small     & 46.0 & 37.3 & 36.1 & 32.5 & 32.1 & 40.5 & 37.4 \\
Llama 4 Scout 17B & 47.2 & 34.1 & 36.9 & 32.5 & 32.5 & 40.5 & 37.3 \\
DeepSeek Coder 33B& 43.7 & 34.9 & 33.7 & 32.1 & 29.8 & 34.5 & 34.8 \\
Qwen 3 8B         & 40.9 & 29.4 & 32.9 & 23.8 & 23.8 & 28.2 & 29.8 \\
Llama 3.1 8B      & 34.5 & 26.2 & 28.2 & 26.6 & 25.0 & 16.7 & 26.2 \\
Granite 3.3 8B    & 27.8 & 20.6 & 23.8 & 23.8 & 24.2 & 6.3  & 21.1 \\
\bottomrule
\end{tabular}
\end{table*}

We benchmarked 16 state-of-the-art models across all 6 supported dialects. Table~\ref{tab:main_results} presents the first comprehensive view of cross-dialect Text-to-SQL performance.

\begin{figure}[t]
    \centering
\includegraphics[width=1\linewidth]{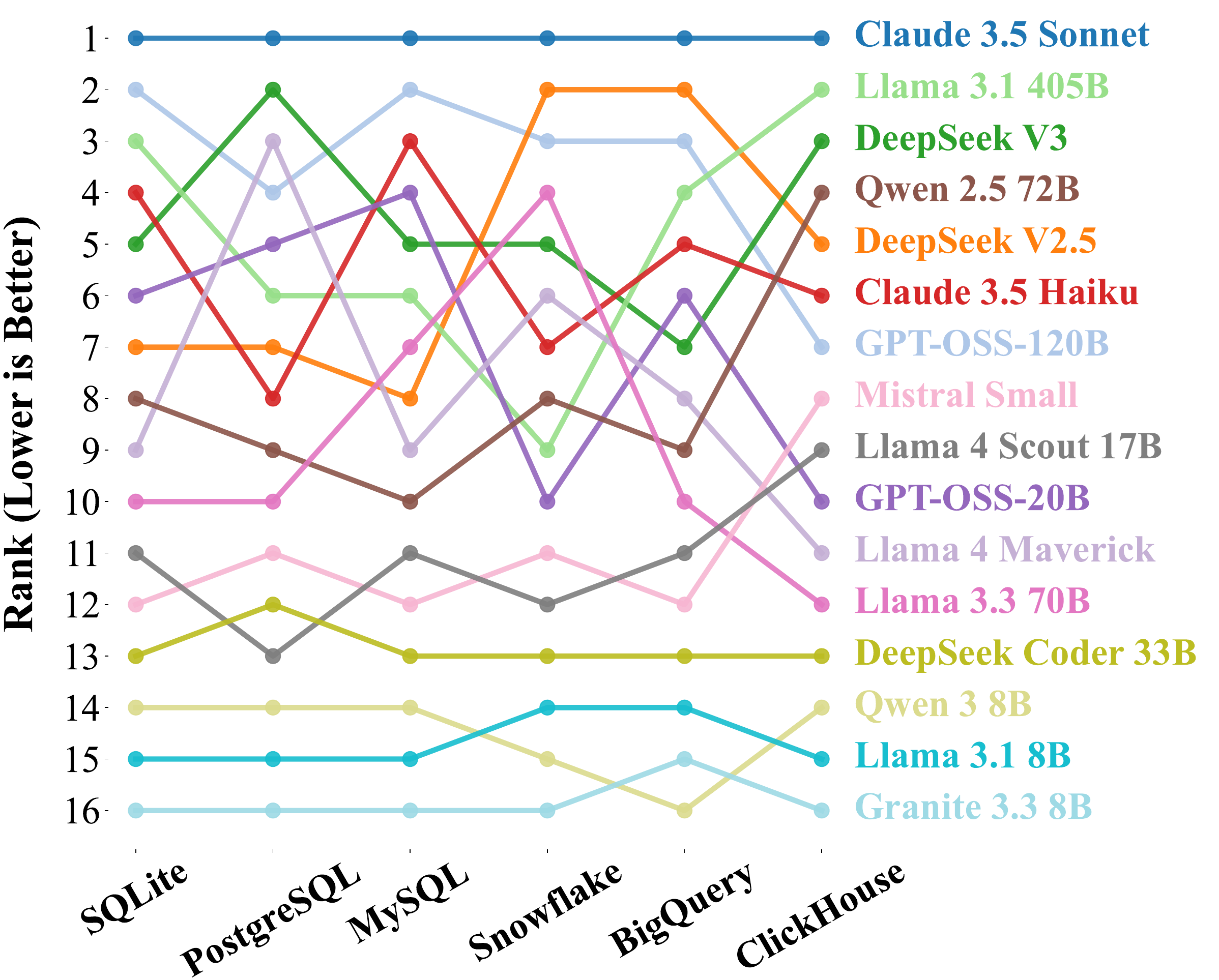}
    \caption{\textbf{Ranking Instability Across Dialects.} The y-axis reports model rank averaged over three datasets, as measured by \frameworkname  on each SQL dialect (x-axis). Rankings on SQLite (leftmost) differ markedly from those on enterprise dialects, revealing substantial rank reordering and limiting the predictive value of SQLite-based leaderboards (detailed analysis in Table~\ref{tab:naive_proxy_failure}).}
    \label{fig:ranking_instability}
\end{figure}

\paragraph{Dialect Distribution Shift.} \textbf{Models exhibit a systematic 10.1\% accuracy drop when evaluated on enterprise dialects compared to SQLite.} This performance gap is statistically significant (paired t-test, $p < 0.001$), indicating systematic rather than incidental differences. We quantify this using our Dialect Robustness Score (Appendix~\ref{app:robustness_score}), which measures the fraction of SQLite performance retained across target dialects.
\paragraph{Rank Volatility.} Performance on SQLite does not perfectly predict performance elsewhere (Figure~\ref{fig:ranking_instability}). While Claude 3.5 Sonnet maintains dominance across all dialects, other models exhibit significant rank changes—for instance, Llama-3.1-405b ranks 2nd on SQLite but 9th on Snowflake. Model selection based solely on SQLite leaderboards thus risks suboptimal performance in enterprise settings.
\paragraph{Dialect Difficulty.} Snowflake and BigQuery prove most challenging (36.0-36.3\% average accuracy), while PostgreSQL and MySQL form an easier tier (39.5-39.8\%). McNemar's test confirms this stratification is statistically significant ($p<0.01$).

\subsection{Cross-Dialect Logic Failures}
\label{ssec:cognitive_fragility}

To understand why performance degrades on target dialects, we analyze the specific failure modes of cross-dialect regressions. We define \textbf{Gap Errors} as queries where the model produces correct SQL on SQLite but fails on the target dialect, isolating pure cross-dialect degradation independent of inherent query difficulty.

We employed GPT-OSS-120B with structured rubrics to classify gap errors (full prompt in Appendix~\ref{app:error_prompt}). We validated this automated approach by manually labeling 100 randomly sampled gap errors, achieving 90\% agreement between automated and manual classifications. Table~\ref{tab:error_categories} presents the distribution of failure modes.

The dominant failure mode (61.2\%) is \textbf{Logic Degradation}—syntactically valid queries with incorrect semantic reasoning—while Dialect Syntax errors account for only 7.7\% of the generalization gap. Logic errors dominate syntactic errors, indicating that dialect shift affects semantic reasoning beyond mere syntax.

Notably, 11.1\% of gap errors were classified as Evaluation Framework issues—primarily ambiguous gold query semantics rather than framework bugs. Even after excluding these cases, Logic Degradation remains the dominant failure mode at 68.8\% of determinate errors.

\section{Related Work}

\subsection{Cross-Dialect Evaluation} Recent efforts to address the fragmentation of SQL dialects generally fall into three categories:

\paragraph{Manual Curation vs. Aggregation.} \textbf{BIRD} \cite{li2023bird} represents the gold standard for fidelity, employing human experts to transpile queries from SQLite to PostgreSQL and MySQL. However, this manual process is prohibitively expensive and static, effectively freezing the benchmark in time. Conversely, \textbf{Spider 2.0} \cite{lei2025spider20evaluatinglanguage} achieves diversity by aggregating disparate datasets from various engines. While the authors performed a limited pilot study ($N=180$) noting that Snowflake was significantly more challenging than BigQuery \cite{lei2025spider20evaluatinglanguage}, this manual analysis was restricted to a tiny subset of the data. Without an automated migration framework, they could not scale this insight to the full benchmark, resulting in a disjointed landscape where dialect difficulty cannot be systematically isolated from question complexity.

\paragraph{Synthetic Adaptation.} Several works leverage synthetic data or intermediate representations to bridge the gap. \textbf{Dialect-SQL} \cite{shi-etal-2025-dialect} and \textbf{SQL-GEN} \cite{Pourreza2024SQLGENBT} generate training samples for Oracle and BigQuery to validate specific adaptation methods. Similarly, \textbf{CrackSQL} \cite{Zhou2025CrackingSB} and \textbf{MoMQME} \cite{Lin2024MoMQME} introduce dialect-specific embeddings. However, these contributions are primarily modeling techniques rather than systematic evaluation frameworks, and they often lack public, execution-ready benchmarks for external validation. Crucially, because no standardized multi-dialect benchmark previously existed, these works were forced to evaluate on proprietary or non-standard dataset splits, rendering their reported "gap closures" difficult to reproduce or compare against other baselines.

\section{Conclusions}

The transition of Text-to-SQL from academic curiosity to industrial necessity requires a commensurate shift in evaluation methodology. To address this evaluation need, we introduced \frameworkname, a framework that replaces brittle query transpilation with execution-based cross-dialect evaluation. By migrating DB state rather than query logic and employing a normalized comparator to bridge engine-specific formatting differences, we achieve the highest benchmark fidelity to date ($\kappa$=0.72) while maintaining 100\% execution coverage. Critically, our framework supports arbitrary source dialects (not just SQLite), with our codebase demonstrating migrations from MySQL, PostgreSQL, and Snowflake.

Our analysis reveals a critical insight: cross-dialect failures are predominantly logical (61\%) rather than syntactic (8\%), challenging the assumption that dialect adaptation is merely syntax translation. This suggests future work should prioritize semantic robustness over pattern matching. For practitioners, SQLite benchmarks provide insufficient signal for production—dialect-specific evaluation is essential for reliable deployment.

Ultimately, \frameworkname{} redefines the standard for robust Text-to-SQL evaluation. By releasing our framework and 15 multi-dialect evaluation environments, we remove the evaluation bottleneck that has constrained cross-dialect research. Future work can now systematically investigate multi-dialect training strategies, dialect-aware prompting techniques, and architectural modifications that preserve reasoning coherence under syntactic variation. As Text-to-SQL systems continue to mature, \frameworkname{} provides the rigorous evaluation foundation necessary to ensure their reliability in diverse, real-world DB environments.

\pagebreak

\section{Limitations}
\label{sec:limitations}

While \frameworkname{} enables scalable cross-dialect evaluation, our methodology involves specific trade-offs regarding feature scope and performance attribution.

\paragraph{Source-Bias and Feature Intersection.}
Our migration pipeline operates on the intersection of source and target dialect capabilities. Because the underlying questions and schemas originate from SQLite-based benchmarks (Spider, BIRD), they do not require—and thus do not test—features exclusive to enterprise dialects, such as Snowflake's semi-structured \texttt{VARIANT} data handling, PostgreSQL's advanced window functions, or BigQuery's array structs. Consequently, our study measures \textit{dialect portability} (how well a model translates generic logic to specific syntax) rather than \textit{native dialect mastery} (how well a model leverages engine-specific optimizations).

\paragraph{Disentanglement of Knowledge vs. Instruction Following.}
Our zero-shot-with-instructions prompting strategy evaluates the model's end-to-end ability to adapt to a target dialect. We observe that performance degradation often manifests as logical errors (\S\ref{sec:main_results}). However, our current experimental design cannot fully disentangle whether these failures stem from a lack of parametric knowledge (the model does not "know" the dialect) or a failure of instruction following (the model ignores the dialect constraints in the prompt). While we mitigate this by using a standardized, verified instruction set, smaller models with weaker instruction-following capabilities may be disproportionately penalized, regardless of their internal representation of SQL syntax.

\paragraph{Normalization Boundaries.}
Our normalized comparator (\S\ref{ssec:protocol}) is designed to be permissive regarding formatting while strict on semantics. However, certain edge cases in floating-point arithmetic and localized collation (sorting orders) vary deeply at the engine level. While we enforce a strict $\epsilon=10^{-5}$ tolerance and canonical sorting, extremely subtle divergences in how different engines handle \texttt{NULL} ordering or mixed-type comparisons may still yield false negatives, though our manual audit suggests this affects $<1\%$ of queries.

\paragraph{Migration Overhead.}
Unlike static analysis evaluation, \frameworkname{} requires active DB infrastructure. Executing 16 models across 15 environments required significant compute resources for DB hosting and migration. While the \textit{human} cost of evaluation is eliminated, the \textit{computational} cost is non-trivial compared to text-matching metrics.

\bibliography{custom,anthology}

\section{Appendices}

\appendix

\section{Automated Error Classification Protocol} \label{app:error_prompt}

To quantify the "Cognitive Fragility" phenomenon, we performed a root-cause analysis on all samples where the model succeeded on the source dialect (SQLite) but failed on the target dialect. To ensure consistency and eliminate human bias, we utilized a standardized LLM judge (\texttt{gpt-oss-120b}) to classify failures into four mutually exclusive categories.

Crucially, our classification protocol enforces a strict distinction between Syntactic Ignorance and Logical Collapse. Failures arising from type system violations (e.g., performing arithmetic on date objects without casting) are classified as \textit{Filtering/Logic Errors}, as they represent a failure to reason about the fundamental data model of the target environment, rather than mere surface-level syntax differences. The full system instruction used for classification is provided below:

\begin{lstlisting}[breaklines=true, language=python, basicstyle=\tiny\ttfamily] 
DEFAULT_JUDGE_TEMPLATE = """ 
You are a meticulous SQL evaluation analyst. Your task is to analyze a single failed NL-to-SQL prediction and classify the PRIMARY root cause of the failure.

=================================================================
CONTEXT: MULTI-DIALECT SQL EVALUATION (THE "GAP" ANALYSIS)
=================================================================

This evaluation tests whether language models can generalize from SQLite (Academic Standard) to Enterprise Dialects (PostgreSQL, BigQuery, Snowflake, etc.).

CRITICAL CONTEXT: 
1. The model's query SUCCEEDED in SQLite (returned correct results).
2. The exact same query (conceptually) FAILED in the Target Dialect (e.g., PostgreSQL).
3. Your job is to explain WHY the transition caused a failure.

Key fields you will receive:
- `gen_type`: The TARGET SQL dialect (e.g., postgres, bigquery, snowflake).
- `schema`: The database schema in the TARGET dialect.
- `predicted_sql`: The model's generated SQL.
- `gold_sql`: A reference query in SQLite (provided ONLY for intent understanding).
- `question`: The natural language question.
- `pred_error`: Error message from executing predicted_sql (if any).
- `results_equal`: Whether both queries returned the same results (False = failure).

=================================================================
YOUR TASK
=================================================================

1. If `results_equal` is True -> output `null` (no error to classify).
2. Otherwise, determine the PRIMARY root cause.
3. Choose the MOST SPECIFIC category that applies.
4. Follow the DECISION PROCEDURE strictly (it prioritizes Logic errors over Syntax errors).

=================================================================
ERROR CATEGORIES (use these exact names)
=================================================================

-------------------------------------------------------------------------------
CATEGORY 1: schema_linking_error (HALLUCINATION)
-------------------------------------------------------------------------------
The model referenced columns or tables that DO NOT EXIST in the provided target schema.

A) COLUMN/TABLE HALLUCINATION:
   - Referencing a column/table name that doesn't exist.
   - Using a column from the wrong table.
   - NOTE: If the table exists in SQLite but NOT in the Target Schema, this is a schema_linking_error (model failed to read the new schema).

-------------------------------------------------------------------------------
CATEGORY 2: filtering_error (LOGIC FAILURE)
-------------------------------------------------------------------------------
The model wrote syntactically valid SQL (or SQL that failed due to type mismatch in a WHERE clause), but the underlying LOGIC selects the wrong rows.

A) LOGICAL MISMATCH:
   - Wrong comparison operators (`>` vs `>=`).
   - Wrong boolean logic (`AND` vs `OR`).
   - Missing JOINs or wrong JOIN conditions.
   - Constraint/Integrity Violations (e.g., query fails due to NOT NULL constraint).

B) TYPE/VALUE LOGIC ERRORS:
   - Comparing incompatible types (e.g., string '2023' vs integer 2023) IF it represents a failure to understand the data model.
   - SQLite-isms that imply wrong typing (e.g., `date_col + 7` treating date as int).
   - *Note:* If this throws a syntax error in Postgres, it is still a FILTERING error (logic) because the model fundamentally misunderstood the data type.

-------------------------------------------------------------------------------
CATEGORY 3: aggregation_error (GROUPING FAILURE)
-------------------------------------------------------------------------------
The model selected the correct rows, but failed in aggregation, grouping, or ordering.

A) GROUPING/OUTPUT LOGIC:
   - Wrong `GROUP BY` columns (often violates strict modes in Postgres/BigQuery).
   - Missing `GROUP BY` when aggregating.
   - Wrong aggregate function (`COUNT` vs `SUM`).
   - Wrong `ORDER BY` direction or column.

-------------------------------------------------------------------------------
CATEGORY 4: dialect_error (SYNTAX ONLY)
-------------------------------------------------------------------------------
The model's LOGIC (Schema, Filtering, Aggregation) is CORRECT, but it used syntax or functions forbidden in the target dialect.

A) SYNTAX IGNORANCE:
   - Using SQLite functions in Target (e.g., `strftime` in BigQuery).
   - Hallucinated functions that exist nowhere.
   - Wrong quoting (backticks vs double quotes).
   - Wrong casting SYNTAX (e.g., `::int` in MySQL).

B) STRICTNESS VIOLATION (SLOPPINESS):
   - Query would work in SQLite but violates strict typing rules in Target (where the logic is arguably correct, but syntax is too loose).
   - EXCLUDES: GROUP BY violations (Must be classified as Category 3).

=================================================================
DECISION PROCEDURE (STRICT ORDER)
=================================================================

1. CHECK FOR SCHEMA REFERENCE ERRORS
   - Did it invent a table or column?
   -> If yes: `schema_linking_error`

2. CHECK FOR ROW SELECTION/LOGIC ERRORS
   - Did it filter on the wrong logic?
   - Did it try to compare a String to an Int (logic flaw)?
   - Did it mess up the JOINs?
   -> If yes: `filtering_error`

3. CHECK FOR AGGREGATION/OUTPUT ERRORS
   - Did it group by the wrong thing?
   - Did it fail a "Strict Group By" check?
   -> If yes: `aggregation_error`

4. CHECK FOR DIALECT SYNTAX ERRORS
   - Everything else makes sense, but it used `strftime` instead of `EXTRACT`?
   - Everything else makes sense, but it used `"` instead of `` ` ``?
   -> If yes: `dialect_error`

5. CHECK FOR EVALUATION/PROCESS ISSUES (Last Resort)
   - The SQL is flawless, but execution failed? (Timeout/Crash)
   - The SQL is flawless, but `results_equal` is False? (Float/Sort mismatch)
   - The Gold Query is wrong?
   -> If yes: `invalid_evaluation`

=================================================================
OUTPUT FORMAT
=================================================================

If `results_equal` is True:
Output only: null

Otherwise output valid JSON exactly like this:
{{
    "question_id": <int>,
    "category": "<schema_linking_error|filtering_error|aggregation_error|dialect_error|invalid_evaluation>",
    "explanation": "<1-2 sentences explaining the root cause>",
    "evidence": "<Short quote from SQL or error message proving the classification>"
}}
=================================================================
PREDICTION DATA TO ANALYZE
=================================================================

{prediction_json}
"""
\end{lstlisting}

\section{Normalized Comparator Specification} \label{app:normalization}

To ensure rigorous cross-dialect comparison, our Normalized Comparator follows a strict three-stage protocol:

\paragraph{1. Structural Alignment (Bag vs. List Semantics).} 
We explicitly handle the distinction between ordered and unordered result sets. Let $Q_{\text{gold}}$ be the ground truth query and $R_{\text{pred}}, R_{\text{gold}}$ be the execution result sets (rows). 

\begin{itemize} 
    \item \textbf{Ordered Context:} If $Q_{\text{gold}}$ contains an explicit \texttt{ORDER BY} clause, we enforce strict list equality. The comparator validates that $R_{\text{pred}}[i] \approx R_{\text{gold}}[i]$ for all indices $i$, penalizing any deviation in row sequence. 
    \item \textbf{Unordered Context (Bag Semantics):} If no ordering is specified, we treat the outputs as multisets. To enable deterministic comparison, we lexicographically sort both $R_{\text{pred}}$ and $R_{\text{gold}}$ by all columns prior to evaluation. This ensures that valid queries are not penalized for engine-specific default sorting (e.g., PostgreSQL's heap scan order vs. SQLite's B-Tree order). 
\end{itemize}

\paragraph{2. Schema Homogenization.} 
To account for dialect-specific column naming conventions (e.g., case sensitivity), we normalize column headers to lowercase and structurally align the dataframes. If the model predicts a superset of the required columns (e.g., returning an extra ID column), we align $R_{\text{pred}}$ to the schema of $R_{\text{gold}}$, discarding extraneous columns while preserving the required signal.

\paragraph{3. Type-Aware Value Comparison.} 
Finally, we perform cell-wise comparison with type tolerance: 

\begin{itemize} 
    \item \textbf{Numeric Tolerance:} Direct equality checks fail across dialects due to floating-point precision differences (e.g., 64-bit vs. 32-bit floats). We employ a tolerance threshold ($\epsilon=10^{-5}$) using \texttt{numpy.allclose}, allowing for minor driver-level variations while rejecting incorrect values. 
    \item \textbf{String Normalization:} String values are whitespace-stripped to handle padding differences (e.g., \texttt{CHAR} vs \texttt{VARCHAR} output). 
    \item \textbf{Type Looseness:} We permit valid cross-type matches (e.g., \texttt{int64} vs \texttt{uint32}) provided the numerical values fall within the tolerance threshold. 
\end{itemize}

This rigorous normalization ensures that our evaluation metric measures \textit{semantic correctness}—the retrieval of the correct data—rather than overfitting to the implementation details of the underlying engine.

\subsection{Algorithmic Specification}

We provide the complete normalized comparison algorithm below for reproducibility and implementation clarity:

\begin{algorithm}[H]
\caption{Normalized Cross-Dialect Result Comparison}
\label{alg:normalized_comparison}
\begin{algorithmic}[1]
\Require Gold query $y^{src}$, predicted query $\hat{y}^{tgt}$, source DB $\mathcal{S}^{src}$, target DB $\mathcal{S}^{tgt}$
\Ensure Boolean correctness signal: $\texttt{True}$ if semantically equivalent, $\texttt{False}$ otherwise

\State $R^{src} \gets \texttt{Execute}(y^{src}, \mathcal{S}^{src})$ \Comment{Execute gold on source}
\State $R^{tgt} \gets \texttt{Execute}(\hat{y}^{tgt}, \mathcal{S}^{tgt})$ \Comment{Execute prediction on target}

\If{$R^{src} = \emptyset$ \textbf{and} $R^{tgt} = \emptyset$}
    \State \Return $\texttt{True}$ \Comment{Both empty: correct}
\EndIf

\If{$|R^{src}| \neq |R^{tgt}|$}
    \State \Return $\texttt{False}$ \Comment{Row count mismatch}
\EndIf

\State \textbf{// Stage 1: Structural Alignment}
\If{$y^{src}$ contains \texttt{ORDER BY}}
    \State $\textit{preserve\_order} \gets \texttt{True}$
\Else
    \State $R^{src} \gets \texttt{LexSort}(R^{src})$ \Comment{Sort all columns lexicographically}
    \State $R^{tgt} \gets \texttt{LexSort}(R^{tgt})$
    \State $\textit{preserve\_order} \gets \texttt{False}$
\EndIf

\State \textbf{// Stage 2: Schema Homogenization}
\State $R^{src}.\textit{columns} \gets \texttt{Lowercase}(R^{src}.\textit{columns})$
\State $R^{tgt}.\textit{columns} \gets \texttt{Lowercase}(R^{tgt}.\textit{columns})$

\If{$R^{tgt}.\textit{columns} \not\subseteq R^{src}.\textit{columns}$}
    \State \Return $\texttt{False}$ \Comment{Missing required columns}
\EndIf

\State $R^{tgt} \gets R^{tgt}[R^{src}.\textit{columns}]$ \Comment{Align column order, drop extras}

\State \textbf{// Stage 3: Type-Aware Value Comparison}
\For{$i = 1$ \textbf{to} $|R^{src}|$}
    \For{$j = 1$ \textbf{to} $|\textit{columns}|$}
        \State $v^{src} \gets R^{src}[i, j]$
        \State $v^{tgt} \gets R^{tgt}[i, j]$

        \If{$v^{src} = \texttt{NULL}$ \textbf{and} $v^{tgt} = \texttt{NULL}$}
            \State \textbf{continue}
        \EndIf

        \If{\texttt{IsNumeric}($v^{src}$) \textbf{and} \texttt{IsNumeric}($v^{tgt}$)}
            \If{$|v^{src} - v^{tgt}| > \epsilon$} \Comment{$\epsilon = 10^{-5}$}
                \State \Return $\texttt{False}$
            \EndIf
        \ElsIf{\texttt{IsString}($v^{src}$) \textbf{and} \texttt{IsString}($v^{tgt}$)}
            \If{$\texttt{Strip}(v^{src}) \neq \texttt{Strip}(v^{tgt})$}
                \State \Return $\texttt{False}$
            \EndIf
        \Else
            \If{$v^{src} \neq v^{tgt}$} \Comment{Direct equality for other types}
                \State \Return $\texttt{False}$
            \EndIf
        \EndIf
    \EndFor
\EndFor

\State \Return $\texttt{True}$ \Comment{All checks passed}
\end{algorithmic}
\end{algorithm}

\textbf{Implementation Notes:}
\begin{itemize}
    \item \texttt{LexSort} performs multi-column lexicographic sorting with NULL-last semantics
    \item Type inference uses driver-level metadata (e.g., \texttt{numpy.dtype} for numeric detection)
    \item Execution failures (syntax errors, timeouts) are caught and return $\texttt{False}$
    \item The complete implementation with type coercion rules is provided in our codebase
\end{itemize}

\section{Dialect Robustness Score}
\label{app:robustness_score}

To quantify model consistency across dialects, we define the \textbf{Dialect Robustness Score} as follows:

\begin{equation}
\text{Dialect Robustness} = 1 - \frac{\text{Acc}_{\text{SQLite}} - \text{Acc}_{\text{target}}}{\text{Acc}_{\text{SQLite}}}
\end{equation}

where:
\begin{itemize}
    \item $\text{Acc}_{\text{SQLite}}$ is the model's accuracy on the SQLite (source) dialect
    \item $\text{Acc}_{\text{target}}$ is the model's average accuracy across the five enterprise target dialects (PostgreSQL, MySQL, Snowflake, BigQuery, ClickHouse)
\end{itemize}

\textbf{Interpretation:}
\begin{itemize}
    \item A score of \textbf{1.0} indicates perfect robustness—the model maintains its SQLite performance across all target dialects (no degradation)
    \item A score of \textbf{0.8} indicates the model retains 80\% of its SQLite performance on average
    \item \textbf{Lower scores} indicate greater dialect-specific fragility and poor generalization
\end{itemize}

This metric provides a single normalized measure of cross-dialect consistency, enabling direct comparison of model robustness independent of absolute performance levels.

\section{Notation Reference}
\label{app:notations}

\begin{table*}[h]
    \centering
    \caption{\textbf{Mathematical Notation Used in Methodology}}
    \begin{tabular}{l l}
        \toprule
        \textbf{Symbol} & \textbf{Definition} \\
        \midrule
        $d$ & SQL dialect (e.g., SQLite, PostgreSQL, MySQL) \\
        $d_{src}$ & Source dialect \\
        $d_{tgt}$ & Target dialect \\
        $\mathcal{B}^d$ & Text-to-SQL benchmark for dialect $d$ \\
        $N$ & Number of examples in benchmark \\
        $q_i$ & Natural language question (example $i$) \\
        $\mathcal{S}_i^d$ & Database state (schema + data) in dialect $d$ \\
        $\mathcal{S}^{src}$ & Database state in source dialect (shorthand for $\mathcal{S}_i^{d_{src}}$) \\
        $\mathcal{S}^{tgt}$ & Database state in target dialect (after migration) \\
        $\mathbb{S}^d$ & Space of all possible database states in dialect $d$ \\
        $y_i^d$ & Ground-truth SQL query for dialect $d$ (gold query) \\
        $y^{src}$ & Gold query in source dialect (shorthand) \\
        $y^{src}_i$ & Gold query in source dialect for example $i$ \\
        $\tilde{y}^{tgt}$ & Synthetic/transpiled query in target dialect (via $\Psi_{tgt}$) \\
        $\hat{y}^{tgt}$ & Model's predicted query in target dialect \\
        $\hat{y}^{tgt}_i$ & Model's predicted query for example $i$ (shorthand) \\
        $p$ & Predicted SQL query from model (alternative notation for $\hat{y}^{tgt}$) \\
        $\Phi(\mathcal{S}^{src}, d_{tgt})$ & Database migration function \\
        $\Psi_{tgt}(y^{src})$ & Query transpilation function to target dialect $d_{tgt}$ \\
        $R^{src}$ & Result set from executing query on source database \\
        $R^{tgt}$ & Result set from executing query on target database \\
        $\epsilon$ & Floating-point comparison tolerance ($10^{-5}$) \\
        $\kappa$ & Cohen's Kappa (agreement metric) \\
        $\rho$ & Spearman rank correlation \\
        $r$ & Pearson linear correlation \\
        \bottomrule
    \end{tabular}
\end{table*}

\end{document}